\documentclass[journal]{IEEEtran}
\usepackage{color}
\usepackage{textcomp,gensymb}
\usepackage{circuitikz}
\usepackage{tikz}
\usetikzlibrary{calc,patterns,decorations.pathmorphing,decorations.markings}
\usetikzlibrary{arrows, shapes, positioning}
\def\blockset#1#2
{\tikzset{
  block/.style = {draw, fill=white, rectangle, minimum height=#1, minimum width=#1, text centered},
  smallblock/.style = {draw, fill=white, rectangle, minimum height=#2, minimum width=#2, text centered},
  narrowblock/.style = {draw, fill=white, rectangle, minimum height=#2, minimum width=2cm, text centered},
  sum/.style = {draw, circle, scale =.5},
  input/.style = {coordinate},
  output/.style = {coordinate},
}
}

\ifCLASSINFOpdf
  \usepackage[]{graphicx}
  \DeclareGraphicsExtensions{.pdf,.jpeg,.png}                       
\else
\fi

\usepackage{amsmath}
\usepackage[caption=false,font=footnotesize]{subfig}
\usepackage{xcolor}
\usepackage{soul}

\begin{document}

\title{
Robotic Wire Arc Additive Manufacturing with Variable Height Layers
}

\author{John Marcotte, Sandipan Mishra, John T.~Wen}

\markboth{Preprint}%
{Marcotte \MakeLowercase{\textit{et al.}}: Robotic Wire Arc Additive Manufacturing with Variable Height Layers}

\maketitle

\begin{abstract}

Robotic wire arc additive manufacturing has been widely adopted due to its high deposition rates and large print volume relative to other metal additive manufacturing processes. For complex geometries, printing with variable height within layers offers the advantage of producing overhangs without the need for support material or geometric decomposition. This approach has been demonstrated for steel using precomputed robot speed profiles to achieve consistent geometric quality.  In contrast, aluminum exhibits a bead geometry that is tightly coupled to the temperature of the previous layer, resulting in significant changes to the height of the deposited material at different points in the part. This paper presents a closed-loop approach to correcting for variations in the height of the deposited material between layers.   We use an IR camera mounted on a separate robot to track the welding flame and estimate the height of deposited material. The robot velocity profile is then updated to account for the error in the previous layer and the nominal planned height profile while factoring in process and system constraints. Implementation of this framework showed significant improvement over the open-loop case and demonstrated robustness to inaccurate model parameters.

\end{abstract}


\IEEEpeerreviewmaketitle

\section{Introduction}

Wire Arc Additive Manufacturing (WAAM) is an additive manufacturing (AM) technology that has become popular due to its high deposition rates and the affordability of equipment relative to other metal additive processes \cite{williamsWireArcAdditive2016}. The first step of using AM to build a part is to decompose the part into layers, often called {\em slices}, that can be laid down one on top of the other to produce the final part geometry \cite{dingAutomatedRoboticArcweldingbased2016}. Typically, these slices are generated by slicing the part with planes parallel to the build surface to create discrete, uniform height layers. One of the challenges that arises is overhanging part geometries, where material needs to be deposited above an empty volume.

\begin{figure}[!t]
\centering
    \subfloat[]{
        \includegraphics[width=0.3\linewidth]{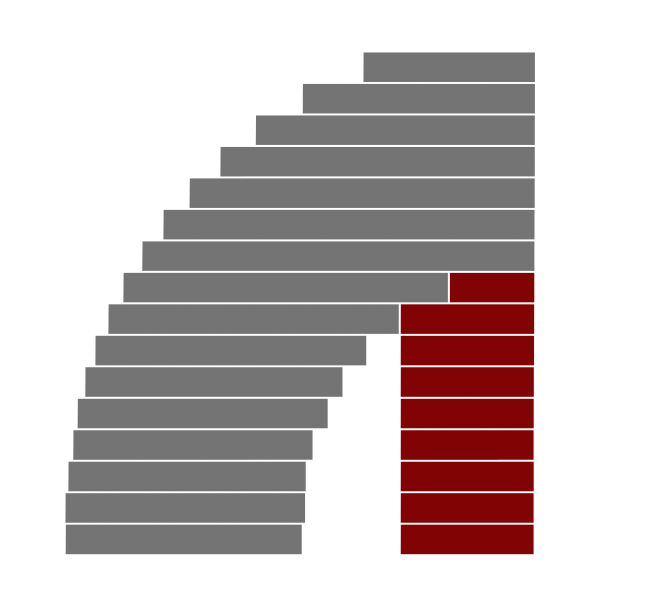}    
        \label{fig:sup_struc}
        }
    \hfil
    \subfloat[]{
        \includegraphics[width=0.3\linewidth]{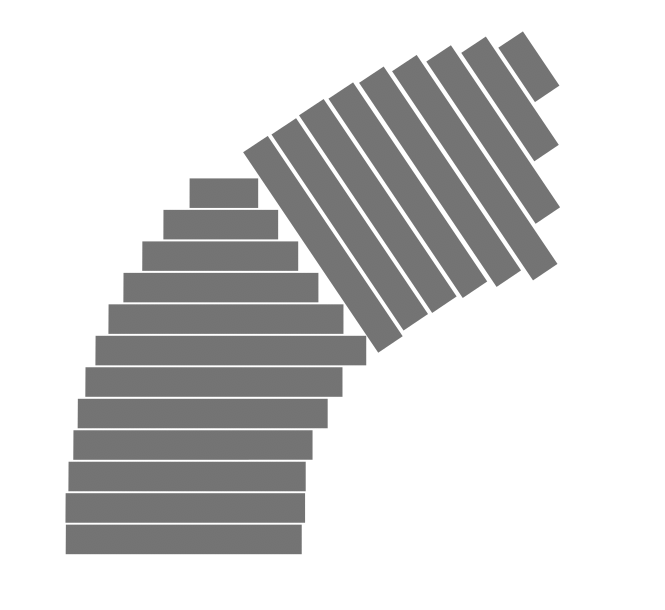}
        \label{fig:geo_decomp}
        }
    \hfil
    \subfloat[]{
        \includegraphics[width=0.3\linewidth]{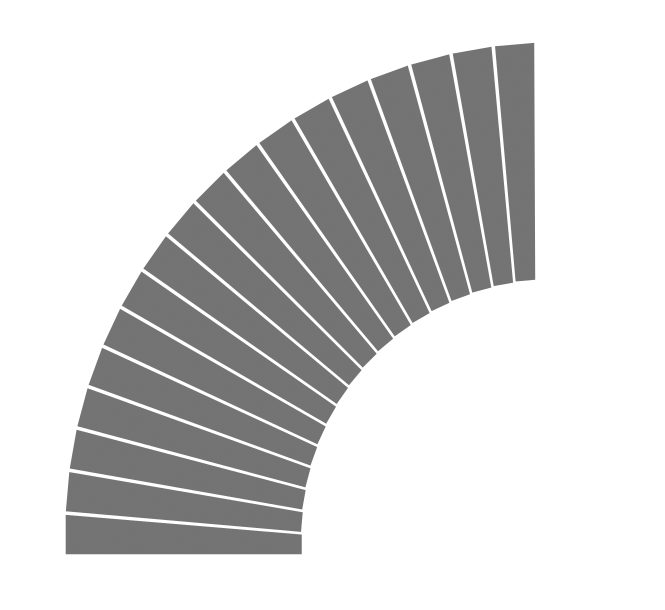}
        \label{fig:ang_layer}
        }
    \caption{Slicing methods for a bent tube using support structures (a), geometric decomposition (b), and non-uniform height layers (c).}
    \label{fig:ang_layer_slice}
\end{figure}

There are several different approaches in the literature to deal with overhanging part geometries. One popular approach, particularly in plastic AM applications, is to lay down extra material to support the overhanging geometry, as illustrated in Fig.~\ref{fig:ang_layer_slice}(a). This method is often used when the part is static, and cannot be reoriented in process. However, it is undesirable for metal AM due to the added difficulty of removing such support material. An alternate approach that has become popular with robotic WAAM systems, where the part can be reoriented in process, is the geometric decomposition approach outlined in~\cite{yuanApplicationMultidirectionalRobotic2019}. This method decomposes the part geometry into several sub-geometries that can be welded by repositioning the part to a more optimal orientation in between (see Fig.~\ref{fig:geo_decomp}). While this method avoids using support material, welding together the sub-geometries results in a rough outer surface, which would require additional machining to achieve the target geometry.

In this paper, we consider the non-uniform height slicing method for overhanging geometries. In this approach, each layer is welded with a non-uniform height to gradually shape the desired geometry. This is illustrated as angled slicing for a bent tube shown in Fig.~\ref{fig:ang_layer}. By decomposing the part into angled slices, we can form overhanging geometry without the need for support material and can continuously weld on the smooth top surface of the previous layers. Others have used a height-velocity model to create angled layer parts using a varying velocity profile to change the height of each layer along a desired profile \cite{rauchNovelMethodAdditive2021,nguyenToolPathPlanning2022,kerberVariableLayerHeights2024}. The process limits that bound attainable heights, and the resulting bounds on the size of parts that can be manufactured using this technique, are demonstrated in \cite{rauchNovelMethodAdditive2021}. All previous works that manufactured parts with angled layers were manufactured without in-process feedback using steel as a feedstock; which demonstrated little error between the expected and measured part.

Aluminum has not been studied for manufacturing parts with variable height layers, possibly due to its lack of geometric consistency as the heat of the part changes. Due to heat accumulation in the part as successive layers are welded, the nominal derived model begins to break down, with the bead geometry changing under the same process parameters. This phenomenon was observed in \cite{dasilvaEffectThermalManagement2021} as they introduced various degrees of cooling to the part and in \cite{ortegaCharacterisation4043Aluminium2019} as they varied the interlayer cooling time. Attempts to replicate the process outlined in \cite{rauchNovelMethodAdditive2021} resulted in a failed part where the distance between the torch and the part grows, ultimately leading to rapid oxidation of the deposited material. As the number of welded layers increased, so did the distance between the torch and the part; ultimately resulting in the weld happening outside of the shielding gas. After welding all of the planned tool paths, the final part hardly resembles the planned geometry. Thus, it is necessary to implement a correction framework for aluminum parts built in this manner to correct deviations in the deposited material as the residual heat in the part increases. 

\section{Problem Statement}
\begin{figure}
    \centering
    \includegraphics[width=\linewidth]{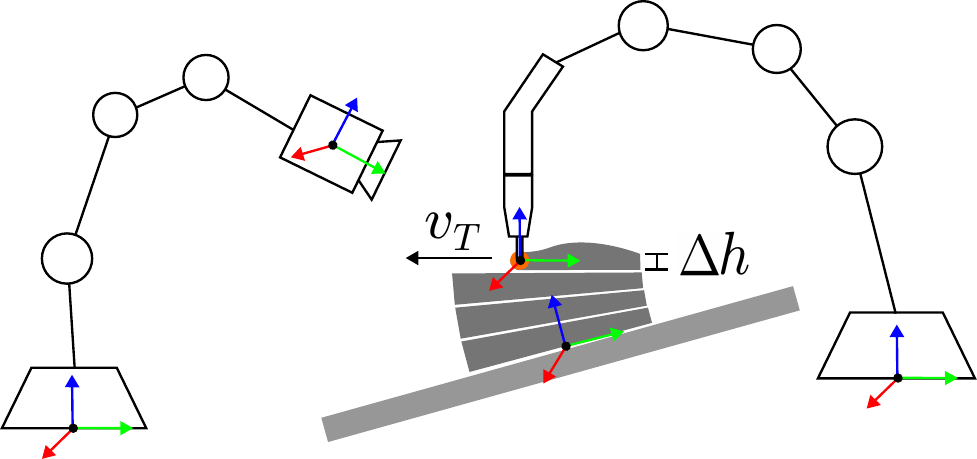}
    \caption{Schematics of the multi-robot WAAM testbed.}
    \label{fig:WAAM-schematics}
\end{figure}

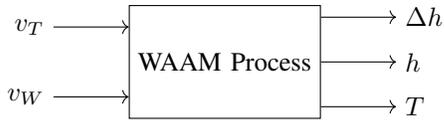
\begin{figure}[h!]
    \centering
\blockset{1.5cm}{1.0cm}
\begin{tikzpicture}[box/.style={draw,text width=1.2cm,align=center}]
\node[block] (P) {WAAM Process};
\node[left=of P.160] (v_T) {$v_T$};
\node[left= of P.200] (v_W) {$v_W$};
\node [right =of P.25] (dh) {$\Delta h$};
\node [right =of P.0] (h) {$h$};
\node [right =of P.335] (T) {$T$};
 \draw[->] (v_T) -- (P.160);
 \draw[->] (v_W) -- (P.200);
 \draw[->] (P.25) -- (dh);
 \draw[->] (P.0) -- (h);
 \draw[->] (P.335) -- (T);
\end{tikzpicture}
    \caption{Process input and output in the WAAM process}
    \label{fig:WAAM_IO}
\end{figure}

The WAAM process involves the torch following the path given by the prescribed slices (at a specified distance above the slice to account for deposition).  
Fig.~\ref{fig:WAAM-schematics} shows a schematic of our WAAM testbed consisting of a welding robot depositing on a trunnion, while a second robot tracks the welding flame. The WAAM robot end effector pose is determined by the torch position and the orientation along the direction of gravity.  The two main parameters that may be controlled during the process are the torch speed, $v_T$, and wire feed rate, $v_W$. The torch speed is the rate at which the torch moves along the surface of the workpiece, and the feed rate is the rate at which the metal wire is fed through the torch.  Deposition of the molten metal results in a change of height from the current height, $h$, by $\Delta h$.  As a lumped parameter approximation, the amount of deposition at a given location is determined by the torch speed, wire feed rate, and temperature (mostly the temperature of the previous layer which affects the solidification rate):
\begin{equation}
    \Delta h = f(v_T,v_W,T).
    \label{eq:f}
\end{equation}
The relevant input/output variables are shown in Fig.~\ref{fig:WAAM_IO}.  

In this work, we use the IR camera to measure the actual height of each layer, $h$.  We fix the feed rate $v_W=v_W^*$, and do not explicitly use the temperature measurement $T$ as the IR camera does not provide reliable temperature measurements for aluminum.  This results in $\Delta h$ as a scalar function of $v_T$: $\Delta h = f(v_T, v_W^*, T)$. The control objective is to use the measured layer height, $h$, to determine $v_T$ so that $h$ tracks the desired height, $h_d$.  The robot motion is specified by the robot program consisting of uniform linear Cartesian motion segments for the torch (using the moveL motion primitive) with specified speed, $v_T$.  The motion program can only be changed between the printing of each layer.  For a scalar variable $x$, denote its value at layer $i$ and segment $k$ by $x_k^{(i)}$, and the stacked variable for the $i$th layer as $\pmb{x^{(i)}}$.  

\section{Methods}

\subsection{Overall Approach}

Our overall control approach is illustrated in Fig.~\ref{fig:WAAM_control}. We first experimentally identify a model for a fixed set feed rate, $v_W^*$, and the temperature during the experiment, $T^*$. The constant feed rate, $v_W^*$, is chosen based on preliminary experiments for consistent welding performance across a range of torch speeds. The entire layer height is denoted as a stacked vector, $\pmb h$, of the average height in each segment.  This is compared with the desired height profile, $\pmb {h_d}$.  The required correction, $\pmb{\Delta h_d}$, is fed into the motion plan to generate the required torch speed profile for the next layer. Note that the actual operating temperature, especially after depositing multiple layers, may be different than the temperature during calibration, $T^*$.  This would affect the accuracy of the model, and consequently the controller performance.

\def\star{^*}
\def\sep{1cm}
\def\halfsep{0.6cm}
\def\distbelow{0.5cm}
\def\lowerdist{1.3cm}
\def\inv{^{-1}}
\begin{figure}[h!]
    \centering
\blockset{1.5cm}{0.5cm}
\begin{tikzpicture}[box/.style={draw,text width=1.2cm,align=center}]
\node[block] (P) {$\stackrel{\mbox{WAAM}}{\mbox{Process}}$};
\node[narrowblock, left=\halfsep of P.160] (finv){Control Plan};
\coordinate [left=\halfsep of finv] (dhd1);
\node[sum, left= of finv] (sum1) {$+/-$};
\coordinate [left=\halfsep of sum1] (coord1){};
\node [input, name=hdinp] (hdinp) {};
\node[above=0.02cm of dhd1] (dh_d) {$\pmb{e}$};
\coordinate [left=0.2cm of sum1] (hd1);
\node[above=0.02cm of hd1] (hd1) {$\pmb{h_d}$};
\coordinate [left=0.2cm of P.160] (v_coord);
\node[above=0.02cm of v_coord] (v) {$\pmb{v_T}$};
\node[left= of P.200] (w) {$v_W=v_W^*$};
\coordinate [right=1.4cm of P.0](h_coord); 
\coordinate [right=1.1cm of P.0] (h_coord1);
\node [above=0cm of h_coord] (h) {$\pmb h$};
\node [right =0.4cm of P.325] (T) {$T$};
\node [right =0.4cm of P.35] (dh){${\Delta h}$};
\coordinate [below=of h_coord1] (h_coord2) {};
\coordinate [right=1.2cm of P.35] (dh_coord);
\coordinate [right=1.4cm of P.35] (dh_coord_out);
 \draw[->] (coord1) --  (sum1);
 \draw[->] (finv) -- (P.160);
 \draw[->] (w) -- (P.200);
 \draw[->] (P.35) -- (dh);
 \draw[->] (P.0) --  (h_coord);
 \draw[->] (P.325) -- (T);
 \draw[-] (h_coord1) -- (h_coord2);
 \draw[->] (h_coord2) -| (sum1);
 \draw[->] (sum1) -- (finv);
\end{tikzpicture}
   \caption{Control Architecture}
    \label{fig:WAAM_control}
\end{figure}
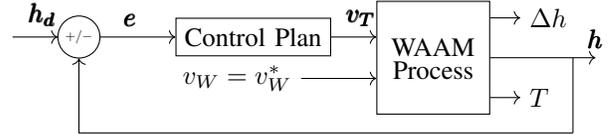
    
\subsection{Process Model}
\label{sec:model}
    The nominal model to relate process parameters to bead geometry was first used in \cite{luMultiRobotScannPrintWire2024}. In this case, the feed rate is fixed, and we are only deriving the relationship between the torch speed and deposited bead height. There are three key assumptions that this model relies on:
    \begin{enumerate}
        \item Volume of material is conserved through the welding process.
        \item The cross-sectional area of the output bead is approximately proportional to the product of the bead height, $\Delta h$, and bead width.
        \item The height and width of the bead are monotonically related. 
    \end{enumerate} 
Under these assumptions, $\Delta h$ is approximately
\begin{equation}
    \Delta h = 
    c v_T^a := \bar f(v_T).
    \label{eq:model1}
\end{equation}    
Taking the logarithm of both sides, we have a linear-in-the-parameter expression:
\begin{equation}
    \ln(\Delta h) = a \ln(v_T) + b
    \label{eq:model2}
\end{equation}
where $a$ and $b$ are constants that depend on $v_W^*$ and $T^*$. Note that since higher torch speed reduces the deposition height at a given location, $\bar f$ is monotonically decreasing (i.e., $a$ is always negative). This also implies that a unique inverse  $\bar f^{-1}(\Delta h)$ is always well defined.

\begin{figure}
    \centering
    \includegraphics[width=\linewidth]{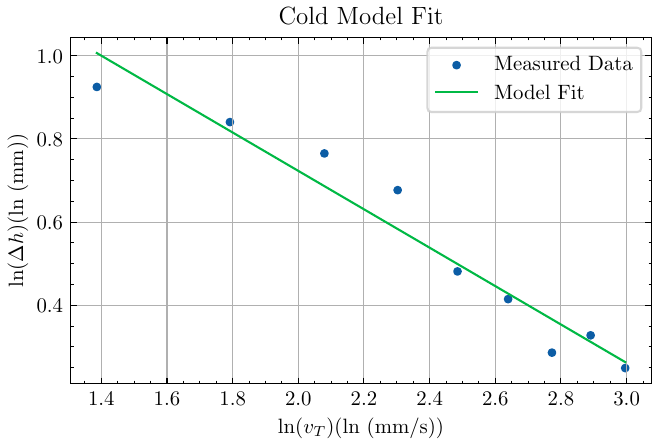}
    \caption{Linear regression of \(\ln(\Delta h)\) and $\ln(v_T)$ based on prior calibration data \cite{luMultiRobotScannPrintWire2024}.}
    \label{fig:regression}
\end{figure}

To identify the model, we adjust the torch speed, $v_T$, and measure the resulting $\Delta h$ under the fixed feed rate.  Estimates of $(a,b)$ may then be obtained through linear regression.  Temperature $T^*$ is difficult to control or measure, but we conducted the calibration experiment under the same condition so the thermal condition is approximately the same.  The regression result is shown in Fig.~\ref{fig:regression} \cite{luMultiRobotScannPrintWire2024}.

\begin{figure}
        \centering
        \includegraphics[width=\linewidth]{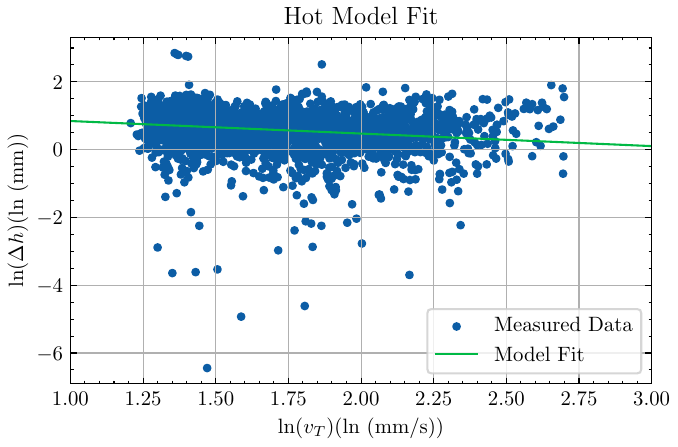}
        \caption{Linear regression of \(\ln(\Delta h)\) and $\ln(v_T)$ based on experimental data.}
        \label{fig:coeff_update}
    \end{figure}

    The primary issue that this work aims to correct is deviations from the model due to thermal disturbances with the part. The thermal conditions under which the data used to fit the model is gathered will change the relationship between the process parameters and the bead geometry. The initial model, which will be referred to as the {\em cold model}, was fit to data where the part was allowed to cool completely between each deposited weld bead.  We denote this model by $\bar f_{cold}$. 
    However, when this model was used to plan and manufacture a part without a large delay between layers, it was found that the model did not fit the data once the part heated up. To account for this, \(v_T\) and \(\Delta h\) were recorded while manufacturing a bent tube without delay, and was fit using a linear regression; showing different values for \(a\) and \(b\) than are seen with $\bar f_{cold}$. The data in this case is dominated by pairs of \(v_T\) and \(\Delta h\) when the part reaches a steady-state temperature. The fit of the new model is shown in Fig.~\ref{fig:coeff_update}, and will be referred to as the {\em hot model}, denoted by $\bar f_{hot}$. The model coefficients for $\bar f_{cold}$ and $\bar f_{hot}$ are shown in Table \ref{tab:mod_coeff}.
    
    \begin{table}[h!]
        \centering
        \begin{tabular}{c|c|c}
            Model & \(a\) & \(b\)\\
            \hline
             $\bar f_{cold}$ & -0.4619 & 1.647 \\
             $\bar f_{hot}$ & -0.3700 & 1.215 
             \vspace{.1in}
        \end{tabular}
        \caption{Model coefficients for the hot and cold models.}
        \label{tab:mod_coeff}
    \end{table}

\subsection{Nominal Process Plan}

\begin{figure}
    \centering
    \includegraphics[width=0.5\linewidth]{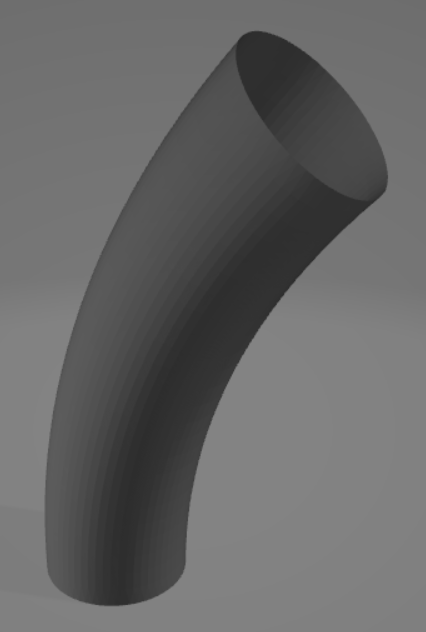}
    \caption{CAD model of the bent tube.}
    \label{fig:tube_cad}
\end{figure}

As a demonstration, we consider the printing of a bent aluminum tube as shown in Fig.~\ref{fig:tube_cad}. The tube has a constant diameter of 50 mm and a radius of 224 mm from the center axis of the tube. The final top surface of the tube was rotated to 45\textdegree{}. The first step is to develop a nominal open-loop plan.  

To start, we will calculate a nominal process plan that can be used to weld a given part using variable height layers. This was done in much the same way as \cite{rauchNovelMethodAdditive2021}, albeit choosing a part that can be manufactured within the radial process bounds without sectioning the part. 

\begin{figure}[h!]
    \centering
    \includegraphics[width=\linewidth]{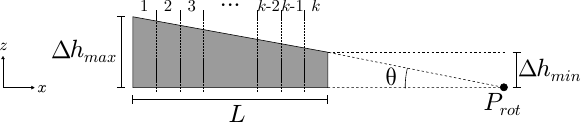}
    \caption{Diagram showing critical dimensions for generating angled layers.}
    \label{fig:layer_slice}
\end{figure}

Given a model of bead height and torch speed for a fixed feed rate, we experimentally derive bounds on the torch speed $[v_{T,\min},v_{T,\max}]$.  For $v_T$ within this range (and $v_W=v_W^*$), the print quality is reasonable and the model, \eqref{eq:model1}, is a good fit to the data. 
The correponding bounds of the deposition heights are $[\Delta h_{\min},\Delta h_{\max}]$, where $\Delta h_{\min}=\bar f(v_{T,\min})$ and $\Delta h_{\max}=\bar f(v_{T,\max})$. Based on the length of the part, \(L\), in the direction perpendicular to the axis of rotation, we can find the rotation center, \(P_{rot}\), and the maximum angular increment, \(\theta\), as shown in Fig.~\ref{fig:layer_slice}. Using the distance from the rotation center, the angle increment, and the process model, we can generate an open-loop plan for each layer, denoted as ${\pmb h_d^{(i)}}$ for the $i$th layer.  The target deposition for the $i$th layer is then 
\begin{equation}
    {\pmb {\Delta h_{nom}^{(i)}}} = {\pmb {h_d^{(i)}}} - {\pmb {h_d^{(i-1)}}}.
    \label{eq:OL_dh}
\end{equation}
Using the calibrated model, \eqref{eq:model1}, we can generate the torch velocity profile:
\begin{equation}
    {\pmb {v_{T,nom}^{(i)}}} = {\bar f}^{-1}(\pmb{\Delta h_{nom}^{(i)}}). 
    \label{eq:OL_vT}
\end{equation}
Once the torch velocity profile is generated, it is converted to coordinated robot motion of the 6-DoF welding robot and a 2-DoF positioner to ensure the deposited layer increment is always along the gravitational acceleration direction \cite{heOpenSourceSoftwareArchitecture2024}. 

\subsection{Height of Layer from IR Camera during welding}

    To measure the height of each layer of deposited material, we use an IR camera mounted to a second 6-DoF manipulator which monitors the weld in situ. The position of the IR camera during welding is shown in figure \ref{fig:weld-cell}.
    The robotically mounted IR camera is calibrated in the world frame based on hand-eye calibration. The positioner and welding robots are also calibrated in the world frame.  The centroid of the bright cluster of the welding arc in the image frame then provides the location of the deposit and may be transformed to the positioner frame for the layer height, $h$. 
    The calibration process and locating the flame through IR imaging is described in \cite{heOpenSourceSoftwareArchitecture2024}. 
    
\subsection{Correction Framework}

\begin{figure}[!h]
        \centering
        \includegraphics[width=\linewidth]{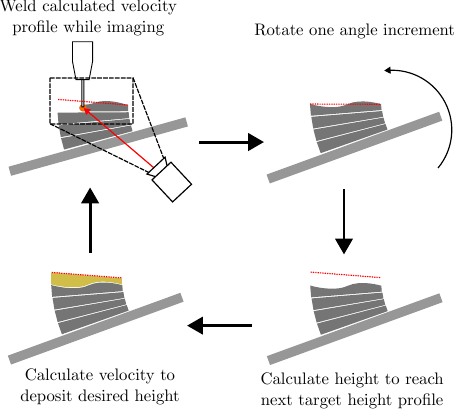}
        \caption{The correction process visualized.}
        \label{fig:cor_proc}
    \end{figure}

The open-loop velocity profile, \eqref{eq:OL_vT}, is based on an imperfect model, \eqref{eq:model1}.  The height measurement enables the calculation of height error in each layer and the adjustment of the torch speed in the subsequent layer to correct for that error.  
Let the error of layer $i-1$ be
\begin{equation}
    {\pmb {e^{(i-1)}}} = {\pmb {h^{(i-1)}}} - {\pmb {h_d^{(i-1)}}}
    \label{eq:error}
\end{equation}
where ${\pmb h^{(i-1)}}$ is from the IR measurement and ${\pmb h_d^{(i-1)}}$ is from the open-loop plan. 
The updated target deposition height for the next layer is then
\begin{equation}
    {\pmb {\Delta h_d^{(i)}}} = {\pmb {\Delta h_{nom}^{(i)}}} -  {\pmb {e^{(i-1)}}}.
    \label{eq:CL_dh}
\end{equation}
This may be used to directly generate the torch speed as in \eqref{eq:OL_vT}.
However, the imperfection of the model could result in torch speeds outside of the acceptable bounds, or large changes in torch speed between neighboring motion segments, which could lead to additional print errors.  
We incorporating these constraints imposing a hard constraint on the torch speed and adding a soft constraint on the speed change.  This results in a quadratic programming problem:
\def\mathbold#1{\pmb{#1}}
\def\layer#1{^{(#1)}}
\def\dh{\Delta h}
\def\norm#1{\left\| {#1} \right\|}
\def\sq{^2}
\begin{align}
& \mathbold{v_T\layer i}=\mbox{arg}\min_{\mathbold{v_T}} \left(
\norm{\mathbold{\dh_d\layer i}- \bar f(\mathbold{v_T})}\sq + \beta \norm{D \mathbold{v_T}}\sq    \right) \nonumber\\
&\quad \mbox{subject to}\quad 
\mathbold{v_{T,\min}\layer i} \prec \mathbold{v_T\layer i} \prec 
\mathbold{v_{T,\max}\layer i}
\label{eq:CL_vT}
\end{align}
where 
$$D = {\scriptscriptstyle
\begin{bmatrix}
    1 & -1 & 0 & \ldots & 0\\
    0 & 1 & -1 &0  &\vdots \\
    & \ddots & \ddots & \ddots &0\\
    & &0 & 1 & -1 
\end{bmatrix}.
}
$$
computes the difference between torch speed in adjacent motion segments. 
The weighting parameter \(\beta\) scales the acceleration penalty and is tunded based on achieved performance. In this work, we choose     
    \begin{equation}
        \beta = \frac{1}{\Delta v_{T,\max}^2}
        \label{eq:beta}
    \end{equation} 
where \(\Delta v_{T,\max}\) is the maximum change in velocity we can tolerate between successive motion segments. 
An illustration of the correction process is shown in Fig.~\ref{fig:cor_proc}.
    
\subsection{Convergence Analysis}
\label{sec:convergence}

The control objective is to select $v_T$ to drive the layer height error, \eqref{eq:error}, close to zero.  
Substituting \eqref{eq:f} into \eqref{eq:error} and then using \eqref{eq:OL_dh}, we have 
\begin{equation}
    \begin{aligned}
     \mathbold{e\layer i} & = 
     \mathbold{h\layer i} - \mathbold{h_d\layer i} \\
& = \mathbold{h\layer {i-1}}  + \mathbold{\Delta h\layer i} - \mathbold{h_d\layer i} \\
& = \mathbold{h\layer {i-1}}  + f(\mathbold{v_T\layer i},v_W^*,T\layer i) - \mathbold{h_d\layer i} \\
 &= \mathbold{h\layer {i-1}}  + f(\mathbold{v_T\layer i},v_W^*,T\layer i) - 
 \mathbold{h_d\layer {i-1}} - \mathbold{\Delta h_{d,nom}\layer {i}  } \\ 
 &=  \mathbold{e\layer {i-1}} + f(\mathbold{v_T\layer i},v_W^*,T\layer i)  - \mathbold{\Delta h_{d,nom}\layer {i}  }.
 \end{aligned}
\end{equation}
\def\e#1{\pmb{e^{({#1})}}}
\def\dhnom#1{\pmb{\Delta h_{nom}^{{(#1)}}}}

For the open-loop control \eqref{eq:OL_vT}, we have
\begin{equation}
    \e i = \e {i-1} + f({\bar f}\inv(\dhnom i),{v_W^*},{T\layer i}) - \dhnom i.
\end{equation}
If we have a perfect model, then $f {\bar f}\inv$ is identity, and the last two terms cancel out.  In this case, the layer height error will remain the same.  Since we start at the base with zero error, the height error will be zero, and we will achieve perfect geometry.  Of course, the model is imperfect, so the error will accumulate and increase as the number of layers increases.

The closed-loop control uses \eqref{eq:CL_dh}. As an approximation of \eqref{eq:CL_vT}, we have
\begin{equation}
    \e i = \e {i-1} + f({\bar f}\inv(\dhnom i-\e{i-1}),{v_W^*},{T\layer i}) - \dhnom i.
\end{equation}
For a perfect model, $f \bar f = I$, and $\e i=0$.  This means that any height deviation from one layer will be fully corrected in the next layer.  When the model is imperfect, $I-f {\bar f}\inv$ is bounded, with the bound dependent on the accuracy of $\bar f$ in approximating $f$.  We shall see in Section~\ref{sec:exp} that the experimental results agree with this analysis.

\subsection{Experimental Setup}

\begin{figure}[!h]
            \centering
            \includegraphics[width=\linewidth]{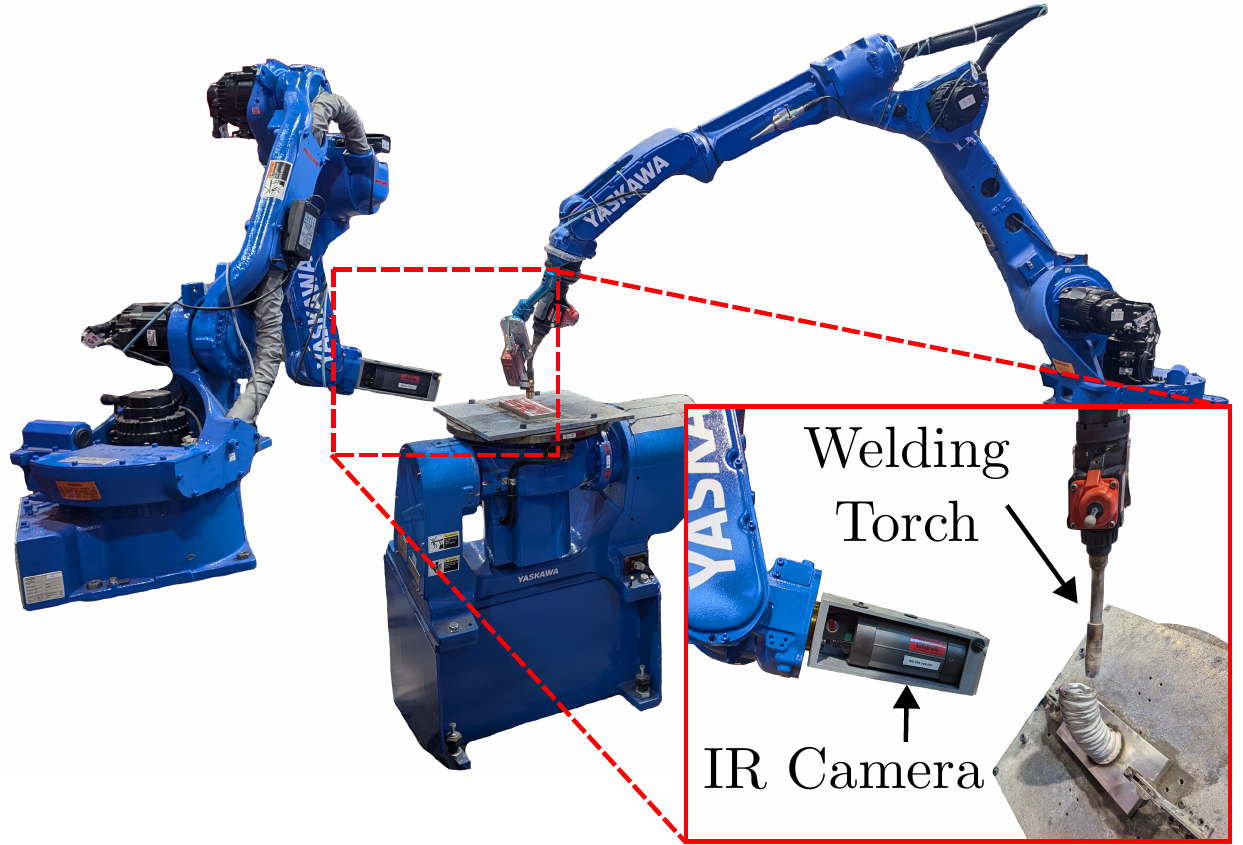}
            \caption{The robotic WAAM testbed.}
            \label{fig:weld-cell}
        \end{figure}

The WAAM testbed at Rensselaer Polytechnic Institute consists of a 6-DoF welding robot with a Fronius weld controller, a 2-DoF positioner, and a 6-DoF monitoring robot equipped with a FLIR IR camera. The Yaskawa Motoman DX200 controller synchronously controls all 14 degrees of freedom.  The details of the testbed is described in  \cite{heOpenSourceSoftwareArchitecture2024}. A photo of the robots in action is shown in Fig.~\ref{fig:weld-cell}. 

\begin{figure}[!h]
            \centering
            \includegraphics[clip, trim = 3cm 0cm 1.5cm 1cm, width=0.5\linewidth]{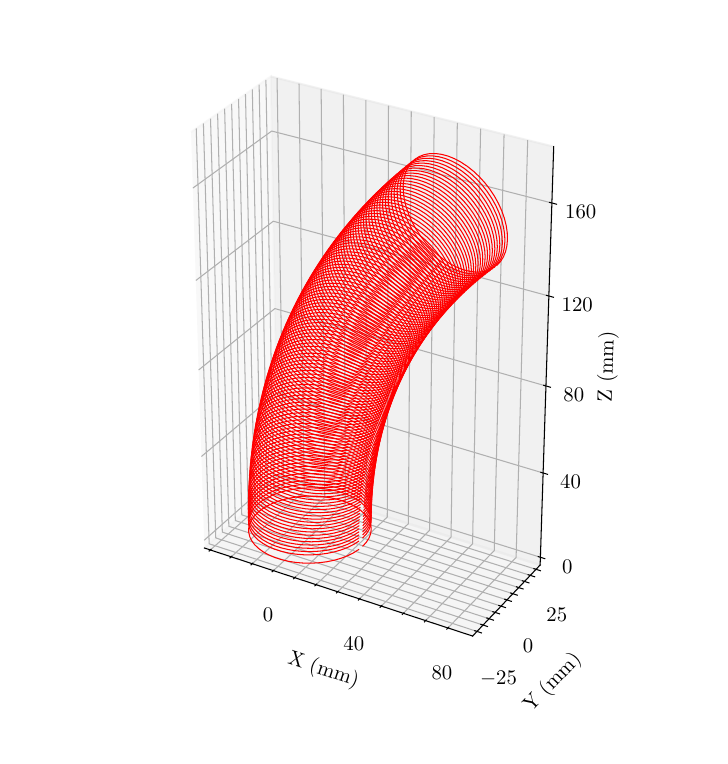}
            \caption{Path plan for the 45\textdegree{} bent tube.}
            \label{fig:tube_plan}
        \end{figure}

For the wire feedstock, we use ER4043 aluminum welding wire. The demonstration part is a 45\textdegree{} bent tube with a diameter of 50 mm, similar to that used in other angled-layer experiments \cite{nguyenToolPathPlanning2022,kerberVariableLayerHeights2024}.   The toolpath planned out for this part is shown in Fig.~\ref{fig:tube_plan}. 
        
\section{Results and Discussion}
\label{sec:exp}

\begin{figure}[h!]
        \centering
        \includegraphics[width=0.6\linewidth]{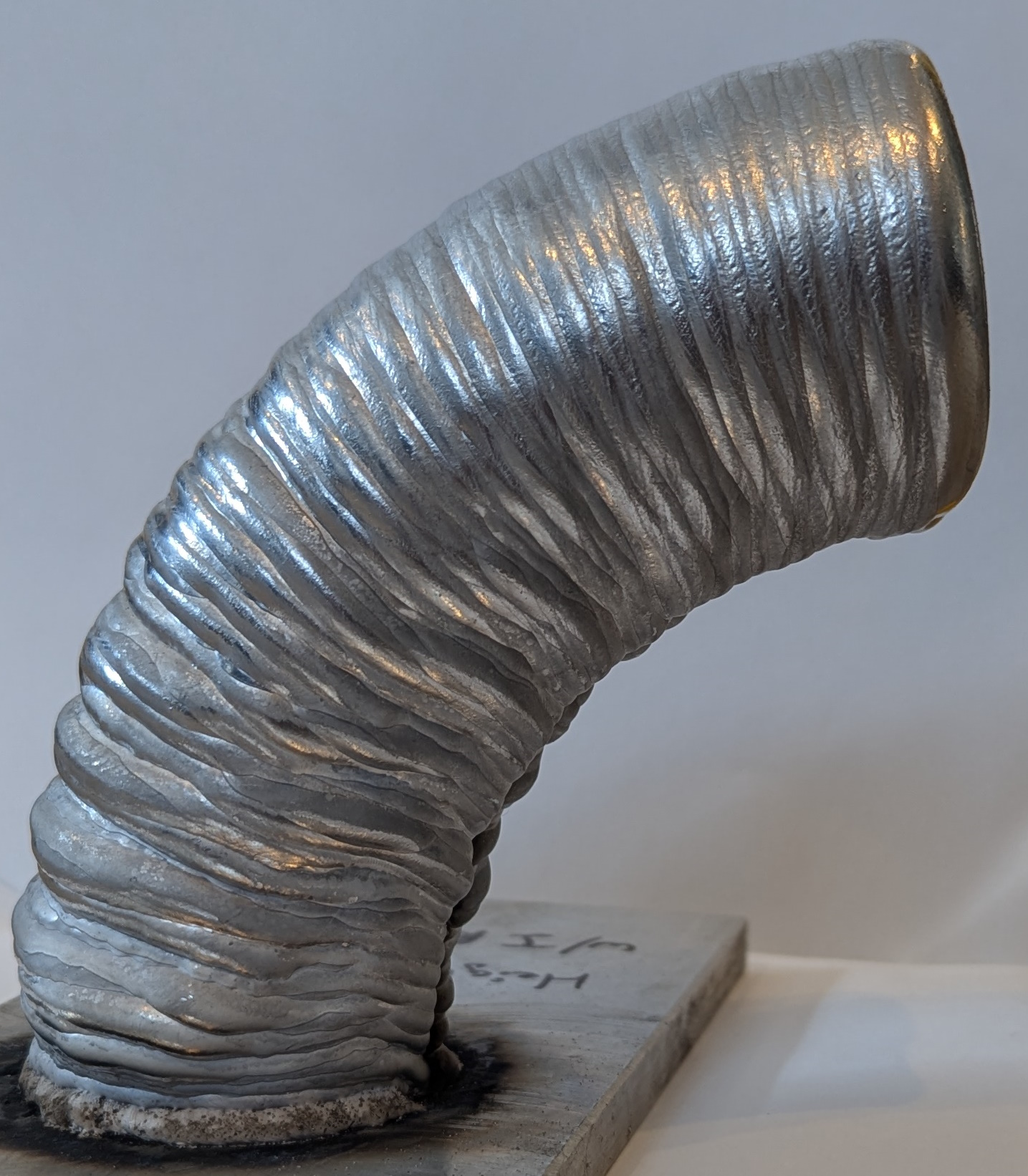}
        \caption{90\textdegree{} bent tube manufactured using the cold model with feedback.}
        \label{fig:cold_90}
    \end{figure}

\begin{figure}[h!]
        \centering
        \includegraphics[width=\linewidth]{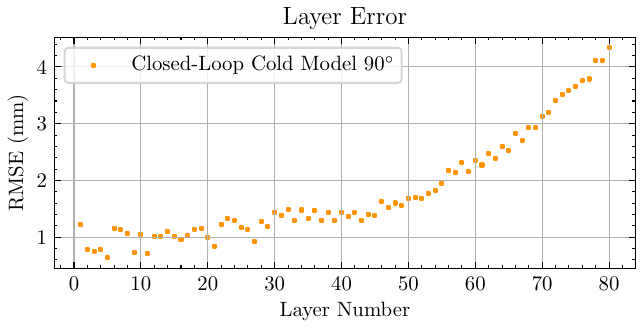}
        \caption{RMSE in each layer of the 90\textdegree{} bent tube manufactured with the cold model. The closed-loop error is almost negligible compared with the open-loop case.}
        \label{fig:cold_90_err}
    \end{figure}

\begin{figure}[h!]
        \centering
        \includegraphics[width=\linewidth]{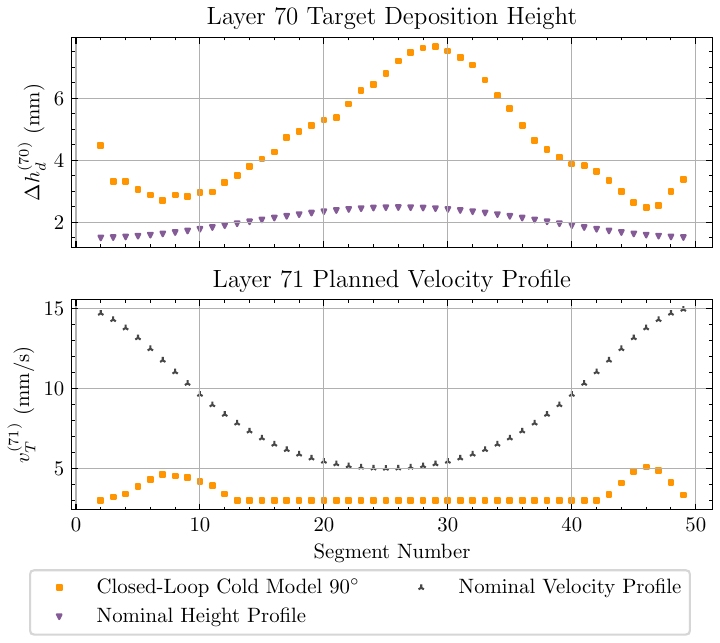}
        \caption{\(\mathbold{\Delta h_{d}^{(70)}}\) and the planned \(\Delta \mathbold{v_T^{(71)}}\) of the 90\textdegree{} bent tube.}
        \label{fig:cold_90_vel_plot}
    \end{figure}

As discussed in Section~\ref{sec:model}, we are considering two models: $\bar f_{cold}$  based on the calibration data when each layer has time to cool down to the ambient temperature and $\bar f_{hot}$ based on the least square estimation of the model parameters $(a,b)$ after multiple layers, without any cool-down period. 
Using the calibrated cold model, $\bar f_{cold}$,  we are able to print a bent tube to 90$^\circ$ as shown in Fig.~\ref{fig:cold_90}. However, the root-mean-square-error of each layer,
    $\norm{\mathbold{e\layer i}}/\sqrt{N}$, where $N$ is the number of motion segments in each layer, shows the error increasing with the layers (Fig.~\ref{fig:cold_90_err}).
    Since the cold model was fit using data with conditions consistent with the first few layers of this part, the correction maintains a low error up to about layer 20. After this point, the nominal process plan has heights that are impossible to achieve within the chosen velocity bounds of the process. This leads to the accumulation of errors as the number of layers increases, even under the closed-loop correction.  This is evident in the error profile in layer 70 and the corresponding commanded torch speed in layer 71, shown in Fig.~\ref{fig:cold_90_vel_plot}.  The nominal deposition in green in the top plot requires a larger height in the middle (to achieve the bent shape)
    resulting in slower torch speed in the bottom plot. The required deposition is chosen sufficiently small to allow for a feasible speed.  In the closed-loop case, the accumulated error (the squares in the top plot) is so large, that the speed is almost constantly at the lower velocity bound (the squares in the bottom plot).  This is still not enough to compensate for the error, resulting in a further increase in error.

To compare the effect of the hot and cold models, we evaluate the following four cases:
\begin{itemize}
    \item {\em Open-Loop, Cold Model} (OC): 
    The velocity profiles are generated by \eqref{eq:OL_vT} with $\bar f=\bar f_{cold}$.
    \item {\em Open-Loop, Hot Model} (OH):
        The velocity profiles are generated by \eqref{eq:OL_vT} with $\bar f=\bar f_{hot}$.
    \item {\em Closed-Loop, Cold Model} (CC): 
        The velocity profiles are generated by \eqref{eq:CL_vT} with $\bar f=\bar f_{cold}$.
    \item {\em Closed-Loop, Hot Model} (CH):       The velocity profiles are generated by \eqref{eq:CL_vT} with $\bar f=\bar f_{hot}$.
\end{itemize}

\begin{figure}[h!]
        \centering
        \subfloat[]{            \includegraphics[width=0.475\linewidth]{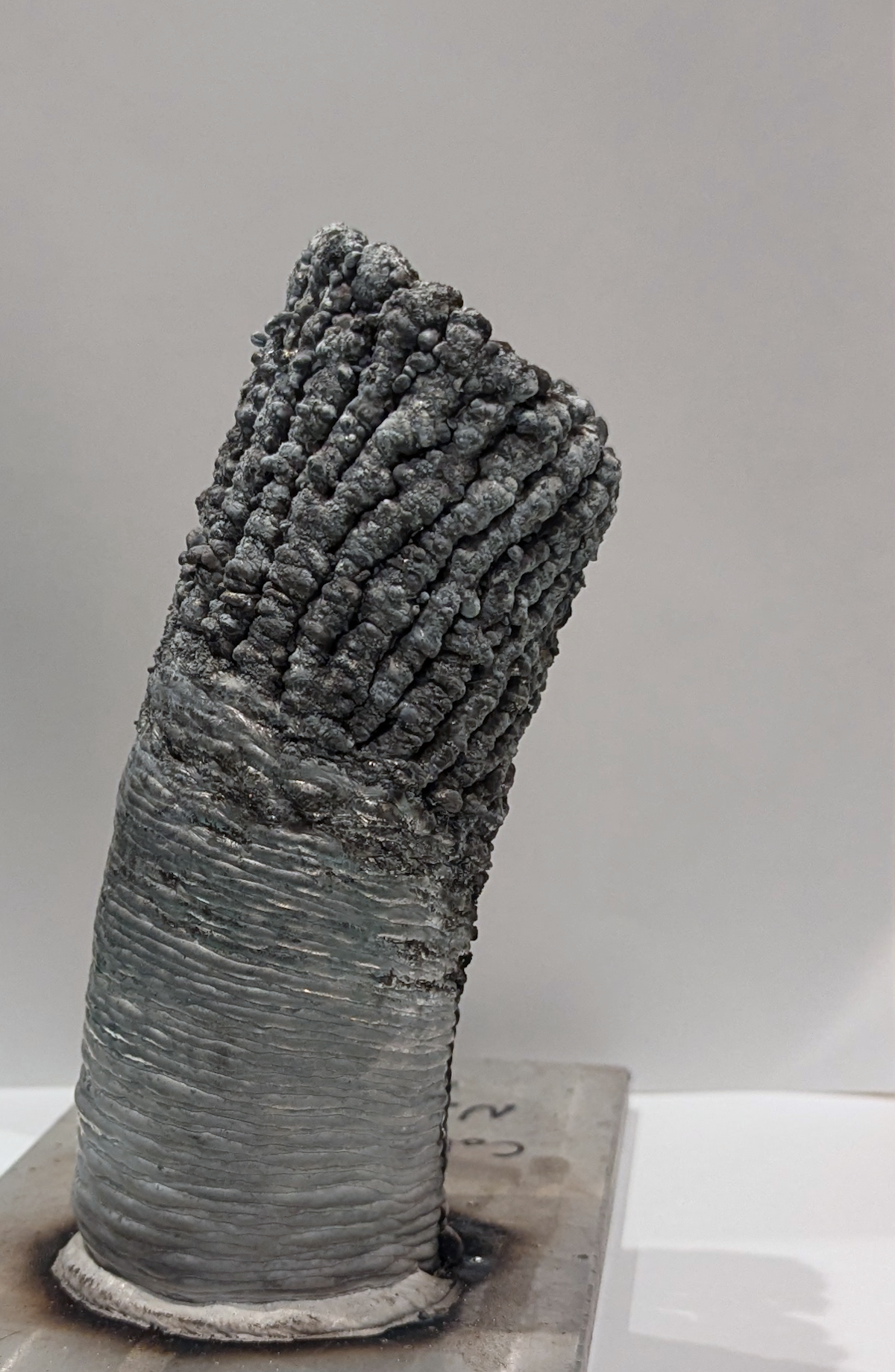}
            \label{fig:ol_cm}
        }
        \subfloat[]{
            \includegraphics[width=0.475\linewidth]{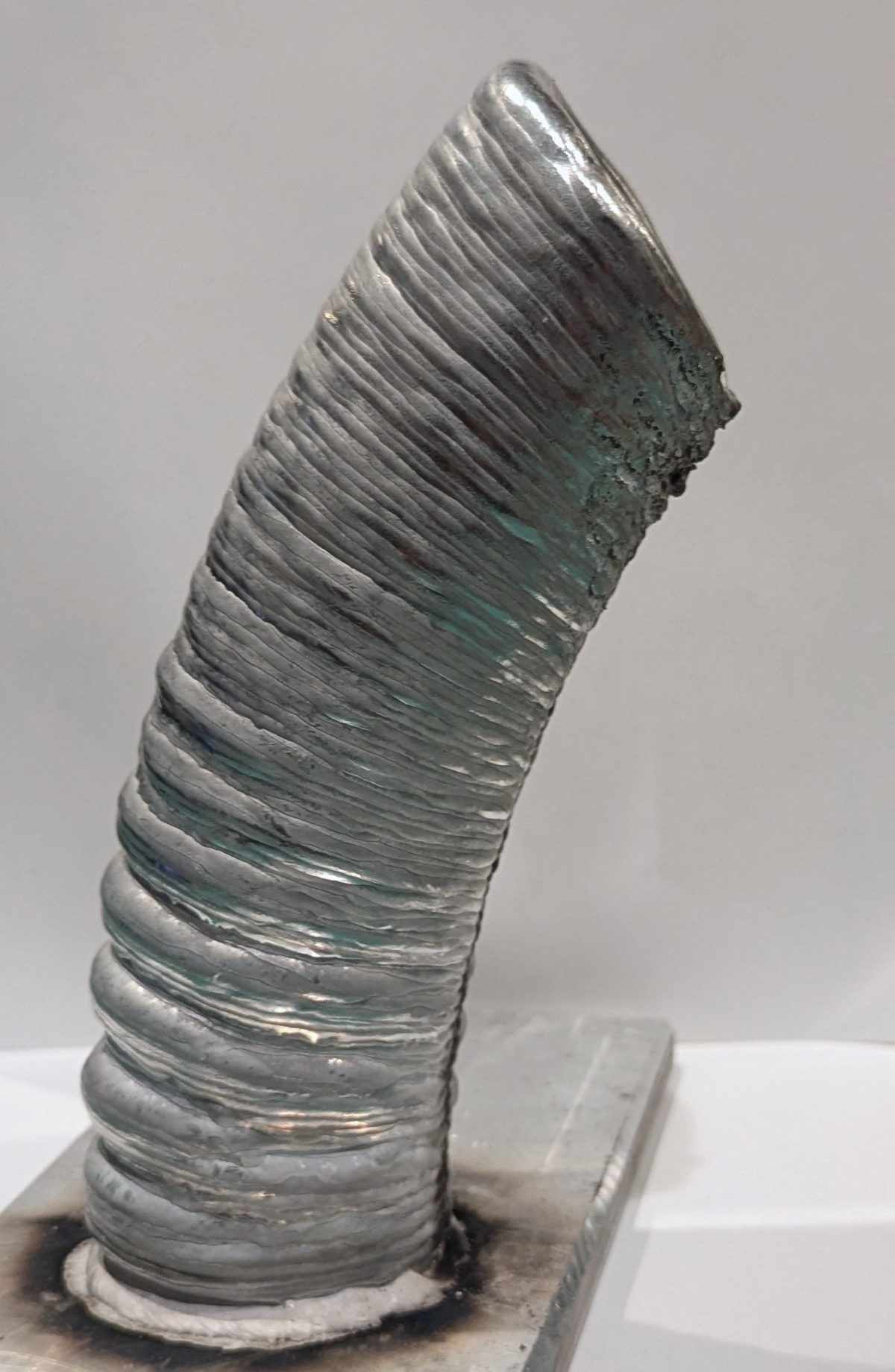}
            \label{fig:ol_hm}
        }
        \vskip\baselineskip
        \subfloat[]{
            \includegraphics[width=0.475\linewidth]{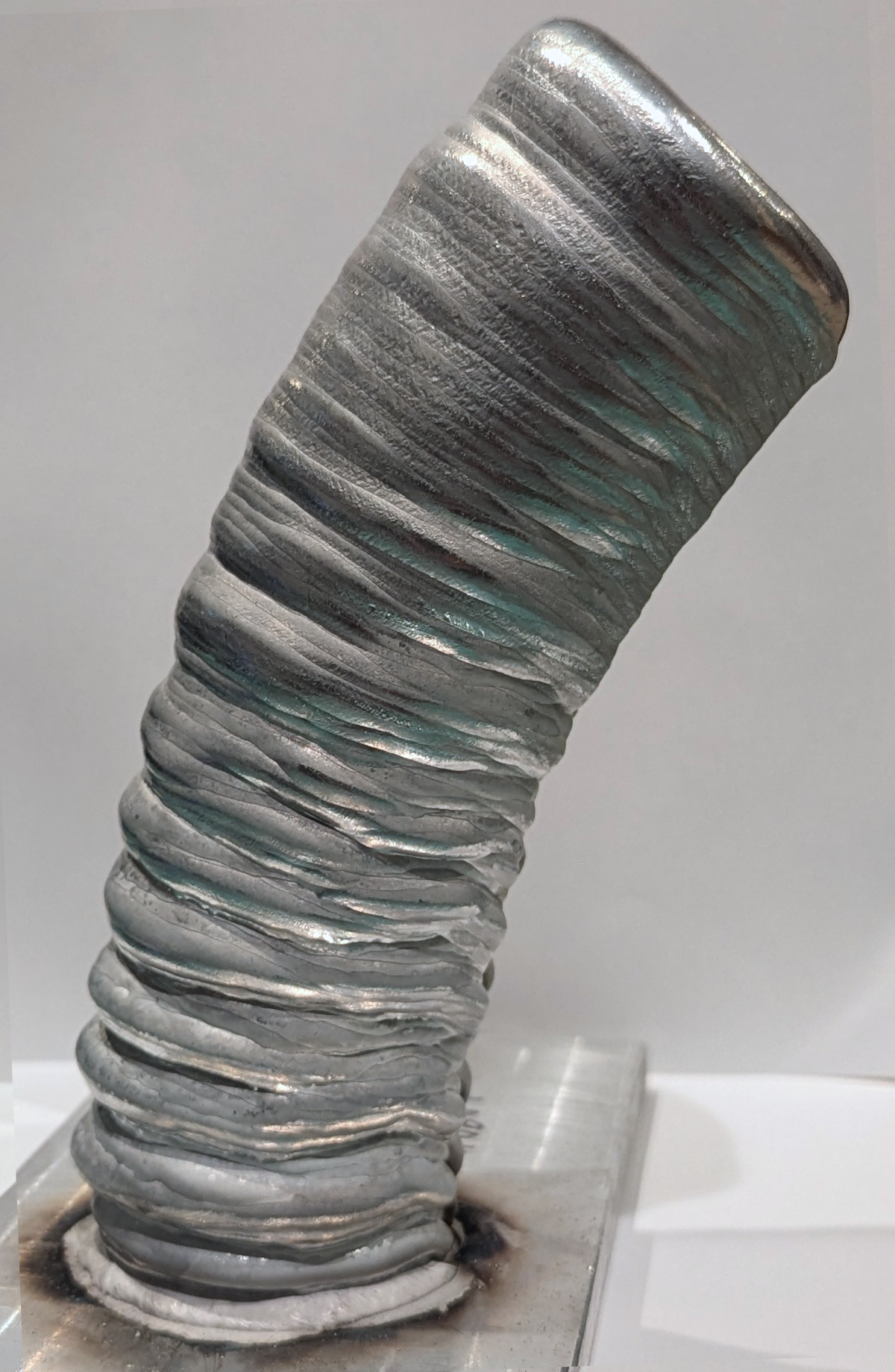}
            \label{fig:cl_cm}
        }
        \subfloat[]{
            \includegraphics[width=0.475\linewidth]{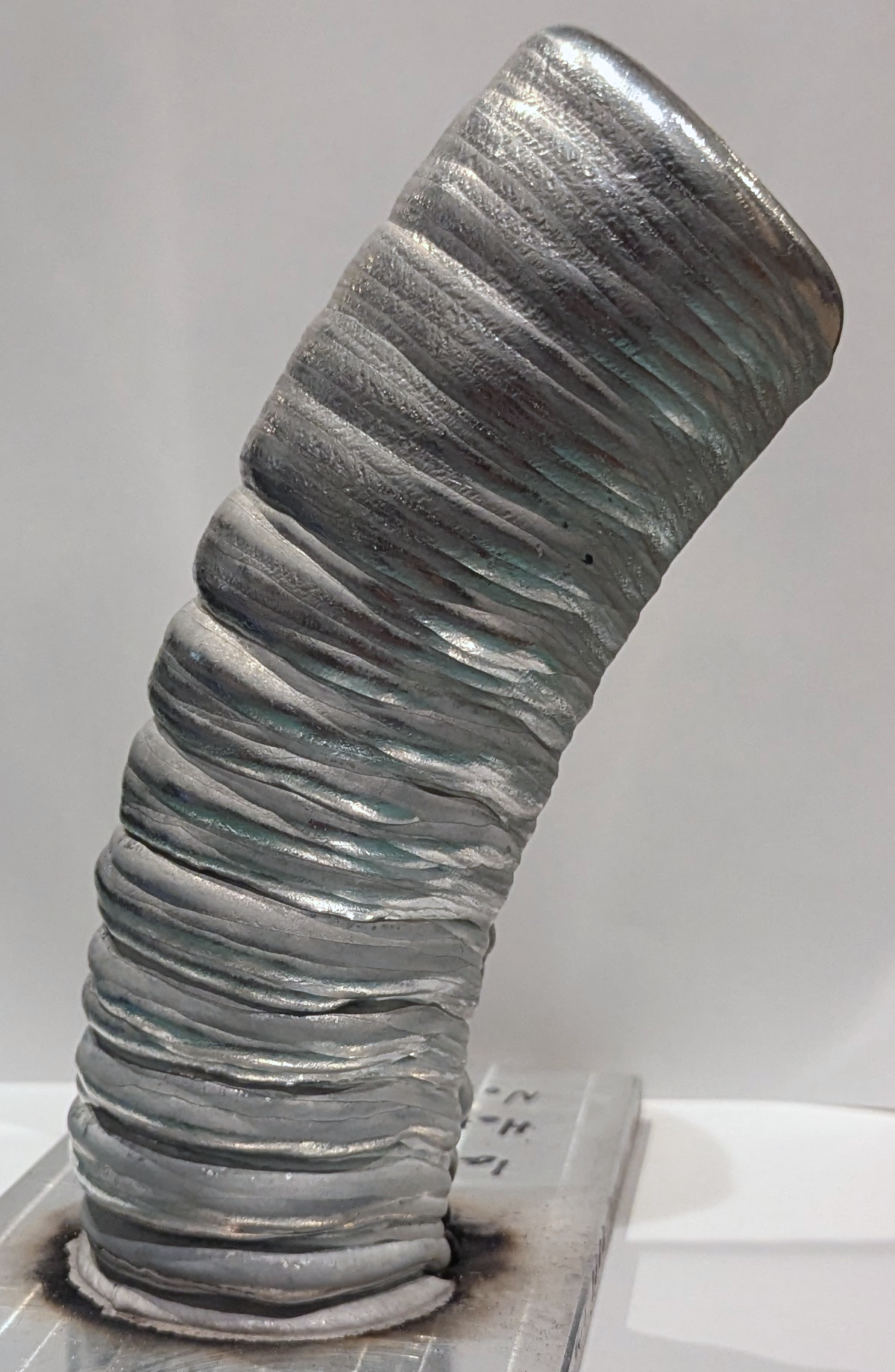}
            \label{fig:cl_hm}
        }
        \caption{The test parts for the open-loop and closed-loop tests with each model: (a) open-loop cold model, (b) open-loop hot model, (c) closed-loop cold model, (d) closed-loop hot model.}
        \label{fig:test_pics}
    \end{figure}

\begin{figure}[h!]
            \centering
            \includegraphics[width=\linewidth]{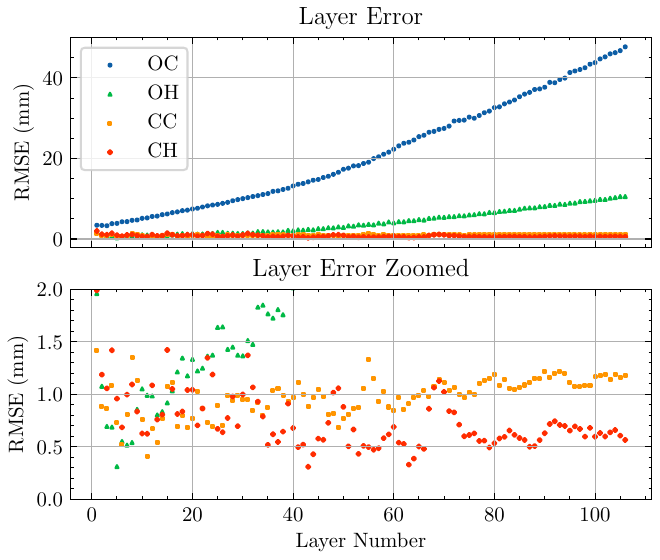}
            \caption{Error comparison between each test case.}
            \label{fig:RMSE}
\end{figure}

Due to bounds placed on the maximum and minimum $v_T$ and the corresponding change in $\Delta h_{max}$ and $\Delta h_{min}$ when using the hot model, the 90\textdegree{} tube was outside the geometric bounds detailed in \cite{rauchNovelMethodAdditive2021}. A new test specimen was created that is inside the geometric bounds for both models to obtain an accurate comparison. 
All four tests are run with the same nominal path plan.  The resulting printed parts are shown in Fig.~\ref{fig:test_pics}.  The maximum layer RMS errors are summarized in Table~\ref{tab:error_result}. The progression of the RMS value of layer error, $\mathbold{e\layer i}$, is shown in Fig.~\ref{fig:RMSE}.

We expect the OC to perform well in lower layers but progressively worse as both the model and print accuracy deteriorate after multiple layers, which is the case as shown in Fig.~\ref{fig:test_pics}(a). Switching to the hot model in the OH case improves the quality but contains a larger geometric error, as shown in Fig.~\ref{fig:test_pics}(b).  With height feedback, case CC allows the complete printing of the bent tube in contrast to the open-loop case, as shown in Fig.~\ref{fig:test_pics}(c).  Using the hot model in the closed-loop further improves the printing accuracy, shown in Fig.~\ref{fig:test_pics}(d).
The OC case is oxidized at higher layers due to the shielding gas being unable to shield the part as the distance between the torch and the part grew. While less apparent, this issue was also present in the OH case.
For both closed-loop tests, oxidation was not present. 
The layer error RMS plot in Fig.~\ref{fig:RMSE} shows the growth of the RMS error in the open-loop cold model case. The error growth improves by using the hot model. Both closed-loop cases avoid the error growth, with the hot model showing better accuracy at higher layers. The experimental results agree qualitatively with the convergence analysis in Section~\ref{sec:convergence}.  The layer height error, $\e i$, grows in the open-loop case with a slower rate of growth under a more accurate model.   In the closed-loop case, the layer error stays within a bounded region, with the size of the bound dependent on the model. 

     \begin{table}[h!]
         \centering
         \begin{tabular}{p{1cm}|p{2cm}|p{1.5cm}|p{1.5cm}}
             \textbf{Model} & \textbf{Feedback} & \textbf{Maximum RMSE (mm)} & \textbf{Final Layer RMSE (mm)}\\
             \hline
             Cold & Open-Loop & 47.73 & 47.73\\
             Hot  & Open-Loop & 10.55 & 10.55\\
             Cold & Closed-Loop & 1.41 & 1.18\\
             Hot & Closed-Loop & 1.99 & 0.57\\
         \end{tabular}
         \vspace{.1in}
         \caption{Error results from all test cases}
         \label{tab:error_result}
     \end{table}

\begin{figure}[h!]
        \centering
        \includegraphics[width=\linewidth]{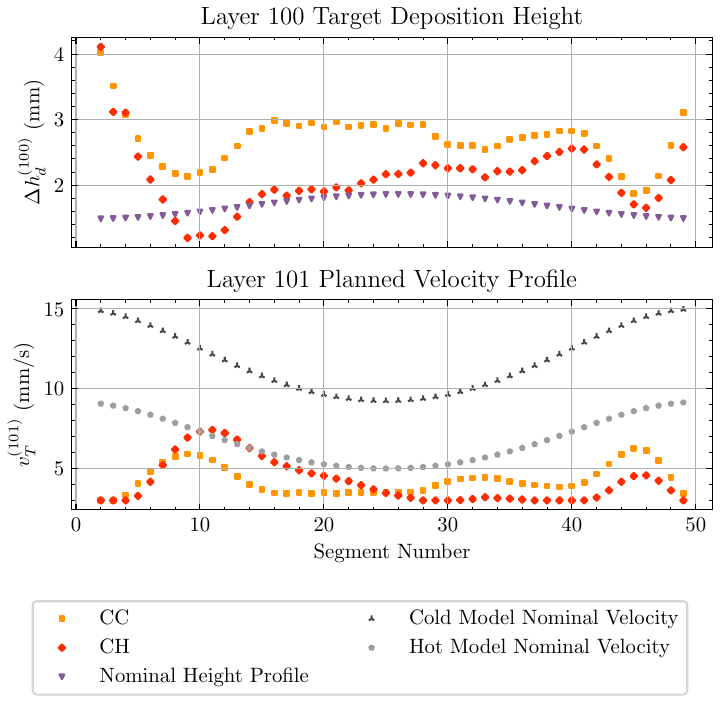}
        \caption{\(\mathbold{\Delta h_{d}^{(100)}}\) and the planned \(\Delta \mathbold{v_T^{(101)}}\) for both closed-loop test cases.}
        \label{fig:100_vel}
    \end{figure}

An example of the target height profile and the resulting height profiles for layer 100 of both closed-loop tests are shown in Fig.~\ref{fig:100_vel} alongside the calculated velocity profile. The output velocity profile is smooth, not reacting to noise oscillations in the desired height profile, demonstrating that the acceleration penalty is reducing the rapid changes of the torch speed within the layer. 

One artifact visible on the final part in several of the tests is the ribbed pattern seen in Fig.~\ref{fig:ol_hm}--\ref{fig:cl_hm}. Since this artifact appears on the hot open-loop and both closed-loop tests, we  deduce that this is not caused by the correction framework, but rather is inherent to the non-uniform layer WAAM process. WAAM often requires a post-process machining step to finish the part, which would remove any visible defects from the surface.   
    
\section{Conclusions}
    In this work, we have proposed and demonstrated a correction framework to ensure accurate material deposition on a nominal path plan for producing angled-layer parts using the WAAM process. The process was analyzed for both the cold and hot models and showed marked improvement over the open-loop tests for both cases. The ability of the framework to successfully manufacture the geometry with varying model parameters demonstrates success even in the presence of an uncertain model.  
    
    Future work will focus on in-process correction rather than layer-by-layer correction.  This would enable parts to be manufactured with one continuous weld bead while correcting the height profile to drive the process back toward the nominal path plan. Introducing the feed rate as another controllable input will allow the width of the deposited weldbead to be controlled similarly. The ribbed pattern on the outside should be investigated and mitigated to minimize necessary post-processing. Outside of the geometric accuracy, the material structure of parts manufactured with this method need to be considered, and should ultimately be factored into the overall correction framework. 

\appendices

\section*{Acknowledgment}
    We would like to thank Honglu He, Chen-Lung Lu, and Jinhan Ren for the WAAM testbed development, the IR camera measurement system, and the generation of the deposition model. 
    
\ifCLASSOPTIONcaptionsoff
  \newpage
\fi

\bibliographystyle{IEEEtran}
\bibliography{bibtex/bibliography_master}

\end{document}